\documentclass[conference,a4paper]{APSIPA2026}
\usepackage{amsmath}
\usepackage{graphicx}
\usepackage{multirow}
\usepackage{threeparttable}
\usepackage{amsmath}
\usepackage{amssymb}
\usepackage{graphicx}
\usepackage{subcaption}
\renewcommand{\figurename}{Figure}

\usepackage{stfloats}
\usepackage[section]{placeins}

\usepackage[backend=biber,style=ieee,]{biblatex}
\usepackage{geometry}
\usepackage{fancyhdr}

\fancypagestyle{firststyle}{
  \fancyhf{}
  \fancyhead[C]{2026 Asia Pacific Signal and Information Processing Association Annual Summit and Conference (APSIPA ASC)}
}

\begin{document}

\title{Topology-Aware Global-Local Mamba Networks for Palm Vein Biometrics}

\author{
\authorblockN{
Zhengxi Wu \authorrefmark{1},
Felix Marattukalam \authorrefmark{1} and 
Waleed H. Abdulla\authorrefmark{1}
}

\authorblockA{
\authorrefmark{1}
University of Auckland, New Zealand \\
E-mail: zwu831@aucklanduni.ac.nz, f.marattukalam,w.abdulla@auckland.ac.nz}

%
}

\maketitle
\thispagestyle{firststyle}
\pagestyle{empty}

\begin{abstract}
Palm-vein recognition is a fine-grained biometric task in which both local vascular texture and the global layout of the vessel tree carry discriminative information, while public datasets remain limited. We propose a topology-aware global-local backbone that combines multi-scale local features, a structure-guided directional stream built on a fixed Sobel-magnitude edge prior, and a four-direction state-space scan global pathway within six Topology-Aware Blocks. A staged gated fusion integrates local, structural, and global representations in that order. On HKPU-NIR, our method achieves $99.13\%$ top-1 accuracy and $0.08\%$ EER with $7.2$~M parameters; on VERA Palm Vein, it achieves $92.42\%$ accuracy and $0.61\%$ EER. Across both datasets it attains the lowest EER among ResNet50, Vim-S, ViT-S, and GLVM at the smallest parameter count, while GLVM remains the strongest in top-1 accuracy and the cheapest in FLOPs. Code is available upon request.
\end{abstract}

\section{Introduction}
Biometric systems play a critical role in modern identity management, and vein-based modalities attract increasing interest because of their subcutaneous vascular patterns, which are observable only under near-infrared (NIR) illumination and are therefore difficult to spoof and stable over time~\cite{sundararajan2018deep}. Among these, palm-vein and finger-vein recognition are particularly suited to high-security applications because of their contactless acquisition, robustness to surface contamination and illumination variation, and requirement for deliberate user participation during capture. However, palm-vein recognition remains challenging due to the thin, low-contrast nature of vascular structures, limited public datasets, and intra-class variation caused by pose, sensor noise, and illumination changes. As a result, effective recognition requires modelling both local vessel texture and global vascular topology.

\subsection{Motivation and Related Work}

Driven by the success of deep learning, convolutional neural networks (CNNs) have largely replaced hand-crafted descriptors in vein biometrics. Prior work has demonstrated strong CNN-based performance for finger-vein and palm-vein recognition under NIR imaging~\cite{hong2017convolutional,das2018convolutional,qin2017deep}. However, stacked convolutions grow their receptive field slowly with depth~\cite{he2016deep}, making it difficult to capture long-range vascular topology without aggressive downsampling that can suppress thin vessel structures.

To improve global context modelling, attention mechanisms such as Squeeze and Excitation, CBAM~\cite{woo2018cbam}, and Transformer-based architectures~\cite{vaswani2017attention} have been explored. While effective at modelling long-range dependencies, self-attention is computationally expensive for high-resolution vein imagery and may dilute sparse vascular patterns.

Selective state-space models (SSMs), particularly Mamba~\cite{gu2023mamba}, provide efficient long-range modelling and have recently been adapted to vision through Vision Mamba (Vim)~\cite{zhu2024vision} and VMamba~\cite{liu2024vmamba}, and to vein recognition through the global-local Vision Mamba GLVM~\cite{qin2026neural}. However, none of these models, including GLVM, explicitly encodes the line-like, direction-sensitive geometry of vascular structures: GLVM uses standard square depthwise kernels in its local branch and additive feature-interaction units for fusion, with no explicit structural prior.

In parallel, medical imaging research has shown that topology-aware priors improve modelling of curvilinear structures. U-Net-based architectures~\cite{marattukalam2020segmentation} and topology-preserving methods such as clDice~\cite{shit2021cldice} highlight the importance of structural continuity in vessel analysis. However, such topological inductive biases have mainly been explored for segmentation rather than biometric recognition, leaving an opportunity to incorporate topology-aware modelling directly into palm-vein classifiers.

\subsection{Proposed Work}
We propose a topology-aware global-local Mamba network for palm-vein biometrics, motivated by the observation that discriminative vein information lies in both local vessel texture and the long-range connectivity of the vascular tree, which are difficult to jointly model using pure CNN, Transformer, or vanilla Vision Mamba architectures.

We evaluate the proposed network on two public palm-vein benchmarks, VERA Palm Vein~\cite{tome2015palm} and HKPU-NIR~\cite{zhang2010online}, under a unified training protocol. Compared with ResNet50~\cite{he2016deep}, Vim-S~\cite{zhu2024vision}, ViT-S~\cite{dosovitskiy2021image}, and GLVM~\cite{qin2026neural}, our architecture achieves the lowest equal error rate (EER) on both datasets with the smallest parameter count, although GLVM retains an edge on top-1 identification accuracy.

Edge operators, strip convolutions, and multi-direction state-space scans are each established in isolation. What this paper contributes is a specific composition of them for palm vein, and the evidence that this composition suits a verification-oriented vein backbone. (1) We inject an explicit, parameter-free and differentiable structural prior, a Sobel-magnitude edge map computed in feature space and shared across the whole body, supplying a structural inductive bias that CNN, Transformer, vanilla Vision Mamba, and the global-local Vision Mamba GLVM~\cite{qin2026neural} leave implicit. (2) We design a direction-sensitive local block that pairs multi-scale depthwise convolutions with axis-aligned strip convolutions, matched to the line-like, sparse, and direction-rich geometry of palm-vein patterns. (3) We fuse in stages, merging the two local streams before the global one so that the four-direction scan operates on a representation that already carries the structural cue. On both benchmarks the resulting backbone reaches the lowest EER of the compared models at the smallest parameter count.

\section{Methodology}

\subsection{Overall Framework}
The proposed topology-aware global-local Mamba network maps a single-channel near-infrared palm-vein image $x \in \mathbb{R}^{1 \times H \times W}$ to a class-logit vector $\hat{y} \in \mathbb{R}^{C}$ for closed-set identification, and exposes a $D$-dimensional descriptor $f \in \mathbb{R}^{D}$ used for verification by cosine similarity.

\begin{figure*}[t!]
\centering
\includegraphics[width=\textwidth]{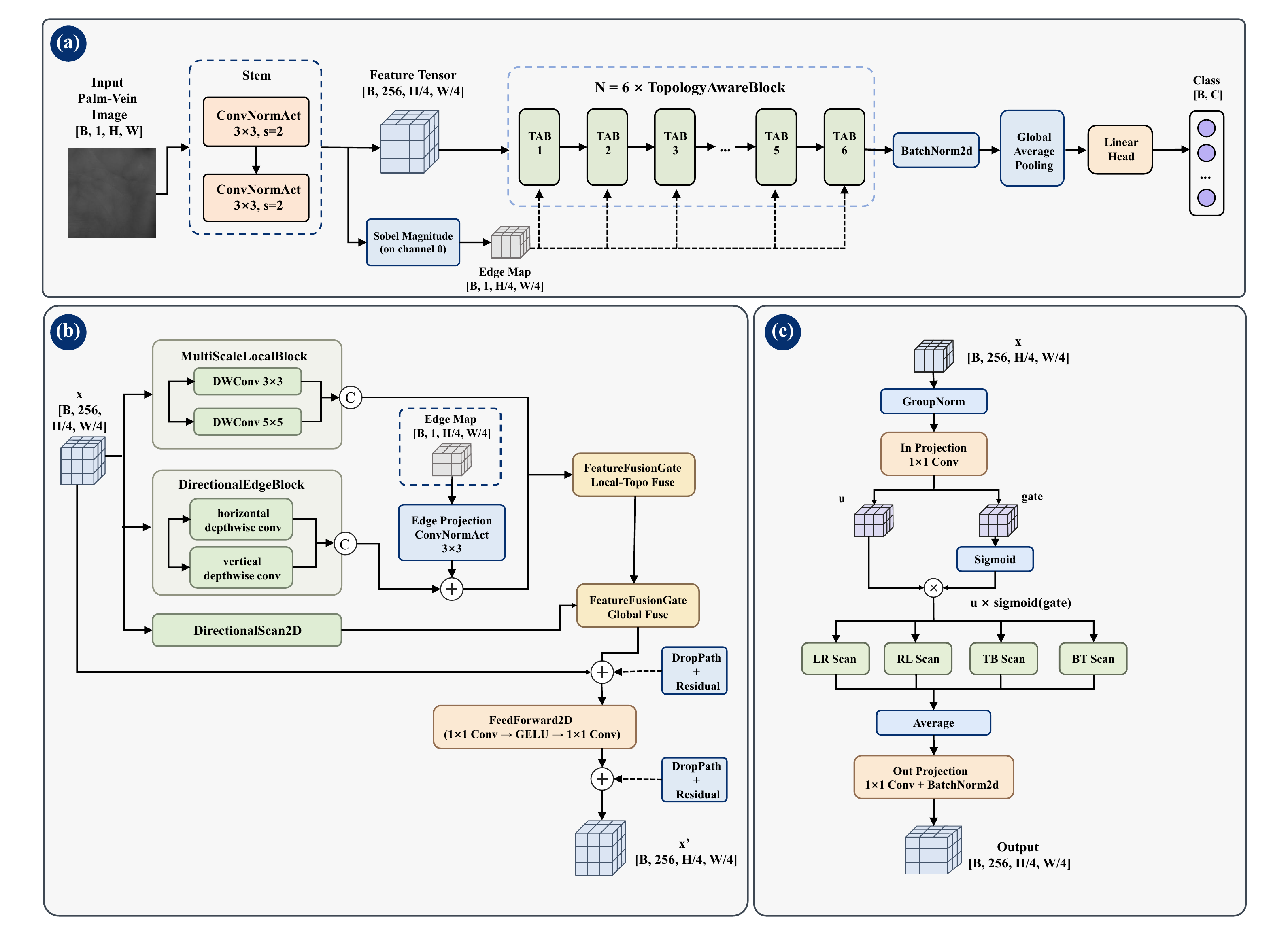}
\caption{Proposed Architecture (a) Overall Pipeline, (b) Topology-Aware Block, and (c) Directional Scan 2D Branch.}
\label{fig:framework}
\end{figure*}

Throughout the paper, $H = W = 128$ and $D = 256$. The network follows the stem-body-head pipeline of Figure~\ref{fig:framework}. Two strided $3 \times 3$ Conv-BatchNorm-SiLU layers reduce the resolution from $128 \times 128$ to $32 \times 32$ and widen the channels $1 \to D/2 \to D$. The body is a stack of $N = 6$ identical Topology-Aware Blocks at constant $32 \times 32$ resolution and constant width $D$. The head applies a final BatchNorm, global average pooling, and a linear classifier to $C$ logits, and the pre-classifier vector is exposed through a separate feature-extraction interface.

Each Topology-Aware Block takes a feature map $z \in \mathbb{R}^{D \times 32 \times 32}$ and preserves its spatial dimensions through three parallel branches and two stages of gated fusion. A multi-scale depthwise convolution branch extracts fine vessel texture; a depthwise strip convolution branch, augmented by an injected Sobel-magnitude structural prior, encodes vessel direction; and, following VMamba~\cite{liu2024vmamba}, a four-direction state-space scan supplies global context by traversing the grid horizontally and vertically in both forward and reverse directions. The first fusion stage combines the multi-scale and structure-augmented directional features into a structurally aware local representation, and the second integrates that representation with the global branch. The fused feature updates $z$ through a residual connection, followed by a pointwise feed-forward network with a second residual update; both residual paths use stochastic depth, with the drop probability increasing linearly from $0$ to $0.08$ across the six blocks.

\subsection{Structure-Aware Directional Local Encoding}
The first two branches of each block form the structure-aware directional local pathway. Palm-vein patterns are thin, low-contrast curvilinear structures whose discriminative information lies in vessel direction and connectivity, so the pathway pairs an isotropic multi-scale stream for vessel texture with an axis-aligned anisotropic stream for directional response, and augments the latter with the structural prior defined below.

Both streams share one two-branch depthwise template. Writing $\phi$ for the SiLU activation, $\star_d$ for depthwise convolution, and $[\,\cdot\,,\,\cdot\,]$ for channel-wise concatenation,
\begin{equation}
\begin{aligned}
\mathcal{B}(z;\mathbf{K}_a,\mathbf{K}_b) = \phi\Big(
\mathrm{BN}\Big(
\mathbf{W} \Big[
&\,\phi\big(\mathrm{BN}(\mathbf{K}_a \star_d z)\big), \\
&\,\phi\big(\mathrm{BN}(\mathbf{K}_b \star_d z)\big)
\Big]
\Big)
\Big),
\end{aligned}
\end{equation}
where $\mathbf{W}$ is a $1 \times 1$ mixing kernel of shape $D \times 2D$. The multi-scale stream instantiates the template with square kernels, $M(z) = \mathcal{B}(z;\mathbf{K}_{3\times3},\mathbf{K}_{5\times5})$, and the directional stream with strip kernels padded so that the $32 \times 32$ resolution is preserved, $D(z) = \mathcal{B}(z;\mathbf{K}_{1\times5},\mathbf{K}_{5\times1})$; the two mixing kernels are parameterised independently. The square kernels capture isotropic vessel texture at two complementary extents, while each strip kernel integrates along one axis only, so it accumulates evidence along a vessel without averaging in the background on either side, which an isotropic kernel of comparable receptive area would do.

The third piece of the local pathway is the Sobel-magnitude structural prior. Let $u = z[:,:1] \in \mathbb{R}^{1 \times 32 \times 32}$ denote the first channel of the post-stem feature map at the entry of the body. The Sobel operator applies fixed $3 \times 3$ kernels $\mathbf{G}_x$ and $\mathbf{G}_y$ in the horizontal and vertical directions, and the prior is the magnitude of the resulting gradient field,
\begin{equation}
E(u) = \sqrt{(\mathbf{G}_x \star u)^2 + (\mathbf{G}_y \star u)^2 + \epsilon},\qquad \epsilon = 10^{-6}
\end{equation}
which is differentiable and carries no learnable parameters. We call $E(u)$ a structural prior rather than a topological one. It is a first-order gradient magnitude, so it measures edge strength, discards the sign of the gradient direction, and is not a topological invariant. Topology-awareness in this work is a property of the block rather than of the operator: $E(u)$ marks where vessel boundaries lie, the strip kernels of $D(z)$ respond to the direction in which they run, and the four-direction scan of Section~II-C links those responses along a vessel, so it is their combination that exposes the branching layout of the vascular tree to the classifier.

Three choices define how the prior is formed and used. First, the response is taken in feature space rather than on raw pixels. The stem acts as a learnable preprocessing stage, so an operator applied on top of it reads the representation the body actually consumes, responds less to raw sensor noise, and already matches the $32 \times 32$ resolution of the encoder features, so no resampling is needed before fusion. Second, the prior is derived from a single channel. One slice keeps the response sharp, whereas averaging $D$ channels before the operator lets gradient responses of opposite polarity cancel, and averaging $D$ magnitude maps after it would pay $D$ Sobel convolutions in every block. The slice is not arbitrary once training begins, because $E(u)$ is differentiable and the only path from the prior back to the stem runs through channel $0$, so the content of that slice is shaped by training rather than fixed in advance. Third, $E(u)$ is computed once at the entry of the body and shared by all six blocks, which keeps it an input-side reference that does not drift with the representation it is meant to constrain; a prior re-derived at depth $k$ would be a function of the very features it is supposed to inform. Adaptation to the evolving feature space is left to the per-block projection $\Pi: \mathbb{R}^{1 \times 32 \times 32} \to \mathbb{R}^{D \times 32 \times 32}$, a $3 \times 3$ Conv-BatchNorm-SiLU layer parameterised independently in each block, which yields $P = \Pi(E(u))$ and augments the directional stream elementwise,
\begin{equation}
\tilde{D}(z) = D(z) + P.
\end{equation}
Every block therefore receives the same structural cue but embeds it in its own channel basis. Whether a multi-channel aggregation or a layer-wise re-derived prior would serve better, and how a raw-pixel Sobel response or a learnable edge extractor would compare, are questions this design leaves open; Section~IV returns to them.

Both fusion points in a block use the same gated mixer. Given streams $a$ and $b$ in $\mathbb{R}^{D \times 32 \times 32}$, a $1 \times 1$ convolution on their channel-wise concatenation followed by an elementwise sigmoid $\sigma$ produces a per-location, per-channel gate, under which the streams are interpolated and refined by a $1 \times 1$ Conv-BatchNorm-SiLU mixer,
\begin{equation}
\begin{aligned}
g &= \sigma\big(\mathbf{W}_g\,[\,a,\ b\,]\big), \qquad g \in [0,1]^{D \times 32 \times 32},\\
\mathrm{Fuse}(a,b) &= \phi\big(\mathrm{BN}\big(\mathbf{W}_m\,\big(g \odot a + (\mathbf{1} - g) \odot b\big)\big)\big),
\end{aligned}
\label{eq:fuse}
\end{equation}
with $\odot$ the elementwise product. The two fusion points hold independent parameters $(\mathbf{W}_g, \mathbf{W}_m)$, since they combine different kinds of stream. The first produces the structurally aware local representation of the block, $L(z) = \mathrm{Fuse}\big(M(z), \tilde{D}(z)\big)$, which carries the multi-scale isotropic texture, the axis-aligned directional response, and the Sobel-derived structural cue in a single tensor, ready to be paired with the global pathway described next.

\subsection{State-Space Global Modelling and Gated Fusion}
While the local pathway captures directional structure within small neighbourhoods, the third branch of each Topology-Aware Block models long-range dependencies across the full $32 \times 32$ feature grid. Following recent vision state-space models such as Vision Mamba~\cite{zhu2024vision} and VMamba~\cite{liu2024vmamba}, we implement this branch as a state-space scanner with a VMamba-style four-direction two-dimensional cross scan. Unlike the original Mamba~\cite{gu2023mamba}, which uses input-dependent discretisation and projection parameters, we adopt compact learnable per-channel scalars to better suit small- to mid-scale palm-vein datasets. Compared with self-attention, the scan provides linear complexity with respect to spatial positions while naturally propagating information along scan directions, aligning well with the elongated routing patterns of palmar vasculature.

The branch runs in three stages: input gating, four one-dimensional scans, and output projection. Gating applies a GroupNorm with $h = 4$ groups and a $1 \times 1$ convolution that doubles the channel width, splits the result into a content tensor $\bar{u}$ and a gate tensor $\eta$, both in $\mathbb{R}^{D \times 32 \times 32}$, and modulates the former by $\sigma(\eta)$:
\begin{equation}
[\,\bar{u},\ \eta\,] = \mathbf{W}_{\text{in}}\,\mathrm{GN}_h(z),\qquad \bar{u} \leftarrow \bar{u} \odot \sigma(\eta).
\end{equation}
The gated content is then traversed as four sequences, left to right, right to left, top to bottom, and bottom to top. Each is processed by an independent state-space module $S$ that updates a per-channel scalar state along the sequence index $t$,
\begin{equation}
s_t = \sigma(\alpha) \odot s_{t-1} + \sigma(\beta) \odot \bar{u}_t,\qquad y_t = s_t + \gamma \odot \bar{u}_t,
\end{equation}
where $\alpha, \beta, \gamma \in \mathbb{R}^{D}$ are learnable per-channel parameters, $s_0 = \mathbf{0}$, and $\bar{u}_t$ is the $t$-th step of the corresponding ordering. Reverse scans flip $\bar{u}$ before the recurrence and flip the output back. The four outputs are averaged and projected by a $1 \times 1$ Conv-BatchNorm layer,
\begin{equation}
G(z) = \mathrm{BN}\big(\mathbf{W}_{\text{out}}\,\tfrac{1}{4}\big(S_{\to}(\bar{u}) + S_{\leftarrow}(\bar{u}) + S_{\downarrow}(\bar{u}) + S_{\uparrow}(\bar{u})\big)\big),
\end{equation}
where the four scans share the recurrence form but keep independent $(\alpha, \beta, \gamma)$.

The structurally aware local representation $L(z)$ from Section~II-B and the global representation $G(z)$ are then merged by the second fusion point, $F(z) = \mathrm{Fuse}\big(L(z), G(z)\big)$, which reuses the form of~\eqref{eq:fuse} with its own gate $g'$ and mixer. Performing the local-structure fusion before the local-global one ensures that the state-space scan is combined with content that already carries the Sobel-derived structural prior and the axis-aligned directional response, rather than with an unstructured local feature alone. We refer to this ordering as a staged gated fusion.

The Topology-Aware Block closes with two residual updates. The fused branch output $F(z)$ is added back to the block input through the first residual,
\begin{equation}
z' = z + \mathrm{DropPath}\big(F(z)\big),
\end{equation}
after which a pointwise feed-forward network refines $z'$ via the second residual,
\begin{equation}
z'' = z' + \mathrm{DropPath}\big(\mathrm{FFN}(z')\big),
\end{equation}
and $z''$ is taken as the block output. The feed-forward network is implemented as a $1 \times 1$ convolution that expands the channel dimension by an MLP ratio of $r = 2$ to a hidden width of $rD$, a GELU activation, and a second $1 \times 1$ convolution that projects back to $D$. The DropPath probability for each block is the corresponding entry of a linear schedule from $0$ to $0.08$ across the six blocks, and the same drop probability is shared between the two residual branches inside one block.

\section{Experiments and Results}

\subsection{Datasets}
We evaluate the proposed method on two public palm-vein benchmarks: HKPU-NIR and VERA Palm Vein. HKPU-NIR~\cite{zhang2010online} contains $500$ classes captured across two sessions. Following the official session-based protocol, the six session-1 images per class train the model and the six session-2 images split into three validation and three test samples, giving $3000/1500/1500$ images. Grayscale images are resized directly to $128 \times 128$ without ROI extraction.

VERA Palm Vein~\cite{tome2015palm} is a contactless NIR palm-vein dataset. Left and right palms are treated as separate identities, giving $220$ classes. We use the five session-1 images per palm for training, while the five session-2 images are randomly split with a fixed seed into two validation and three test samples, yielding $1100$ training, $440$ validation, and $660$ test images. Official ROI crops are resized to $128 \times 128$ grayscale.

For both datasets, pixel intensities are normalised to $[-1,1]$ and the network operates on a single input channel throughout.

\begin{figure*}[!t]
\centering
\includegraphics[width=\textwidth]{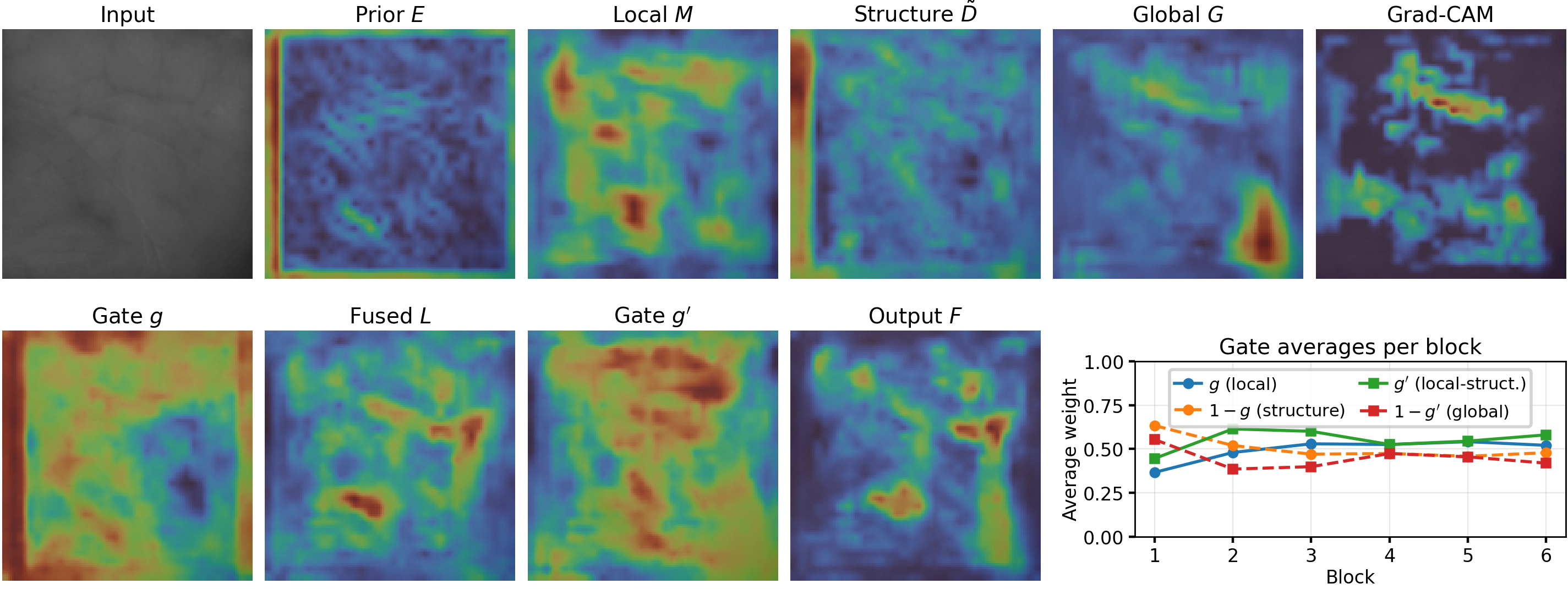}

\vspace{-2mm}

\caption{Branch responses and staged fusion on a VERA Palm Vein test sample.}
\label{fig:branch_gate}
\end{figure*}

\subsection{Implementation Details}
We optimise the model with AdamW using a weight decay of $1 \times 10^{-4}$. We disable weight decay on bias terms, BatchNorm parameters, and the per-channel state-space scalars of the directional Mamba branch. The learning rate follows a five-epoch linear warmup from zero to a peak of $3 \times 10^{-3}$, after which a cosine schedule anneals it to a minimum of $5 \times 10^{-4}$. We train each model for $500$ epochs, select the checkpoint with the highest validation top-1 accuracy, and evaluate that checkpoint once on the held-out test set.

For training-time regularisation we combine label smoothing at factor $0.1$ with Mixup at $\alpha = 0.2$. Geometric augmentation applies a random affine transform with rotation in $[-10^{\circ}, 10^{\circ}]$, translation up to $5\%$ of the image side, and scaling in $[0.9, 1.1]$, followed by photometric jitter on brightness and contrast with strength $0.2$. We omit horizontal flipping on both datasets, because VERA labels left and right palms as separate identities and flipping would create an irrecoverable supervision conflict. The batch size is $128$ on HKPU-NIR and $32$ on VERA. All experiments run on a single NVIDIA A800 (80GB) GPU under PyTorch 2.5.1 with CUDA 12.4.

\subsection{Evaluation Metrics}

We report the proposed method under both the identification and the verification protocols on each dataset. For identification we measure closed-set top-1 accuracy on the test split, treating the dataset as a fixed-class classification problem and reading the predicted class from the $\arg\max$ of the classification head.

For verification we take a $256$-dimensional descriptor from the pre-classifier interface, $L_2$-normalise it, and score every unordered test pair by cosine similarity. Sweeping the decision threshold $\tau$ gives the false-accept rate (FAR) and false-reject rate (FRR), and we report $\mathrm{EER} = \tfrac{1}{2}(\mathrm{FAR}(\tau^{\star}) + \mathrm{FRR}(\tau^{\star}))$ at $\tau^{\star} = \arg\min_{\tau}|\mathrm{FAR}(\tau) - \mathrm{FRR}(\tau)|$.

To characterise model cost we also report the total parameter count (M) and the FLOPs of a single forward pass on a $1 \times 128 \times 128$ input (G).
\vspace{-1.5mm}

\subsection{Comparison with Representative Backbones}
We compare our method against four published baselines: ResNet50~\cite{he2016deep}, Vim-S~\cite{zhu2024vision}, ViT-S~\cite{dosovitskiy2021image}, and the global-local Vision Mamba GLVM~\cite{qin2026neural}. Their numbers are quoted from~\cite{qin2026neural}, which evaluates them on the identical session-based partition, $3000/1500/1500$ images on HKPU-NIR and $1100/440/660$ on VERA, and defines accuracy, EER, parameters, and FLOPs as they are defined here. Our recipe follows the configuration reported in~\cite{qin2026neural} except for the batch size and the peak learning rate, retuned for the proposed architecture. The comparison is thus matched in data and protocol.

Table~\ref{tab:compare} reports both benchmarks. The proposed method attains the lowest verification EER on each ($0.08\%$ on HKPU-NIR and $0.61\%$ on VERA Palm Vein), while GLVM attains the highest top-1 identification accuracy ($99.40\%$ and $95.13\%$). Our model is also the smallest in parameter count ($7.2$~M), but at $7.28$~G its forward pass costs an order of magnitude more than GLVM at $0.59$~G. The reason is architectural: the body never downsamples, so all six blocks run three branches, two gated mixers, and a feed-forward network at the full $32 \times 32 \times 256$ shape. The global pathway accounts for $2.43$~G and the $1 \times 1$ mixers and feed-forward networks for the remaining $4.85$~G, so the cost belongs to the dense multi-branch block rather than to the scan alone. One image takes $26$ to $34$~ms at batch size one on the A800, the recurrence running step by step in PyTorch rather than through a fused kernel. A verification gate can absorb this in exchange for the lowest EER in Table~\ref{tab:compare}; where the inference budget dominates, GLVM remains the better choice.

\begin{table}[t]
\centering
\caption{Comparison on HKPU-NIR and VERA Palm Vein.}
\label{tab:compare}
\small
\setlength{\tabcolsep}{3pt}
\begin{tabular}{lcccccc}
\hline
 & \multicolumn{2}{c}{HKPU-NIR} & \multicolumn{2}{c}{VERA} & \multicolumn{2}{c}{Cost} \\
\cline{2-3}\cline{4-5}\cline{6-7}
Method & ACC & EER & ACC & EER & Params & FLOPs \\
       & (\%) & (\%) & (\%) & (\%) & (M) & (G) \\
\hline
ResNet50~\cite{he2016deep}         & 98.93 & 0.39 & 89.55 & 2.48 & 24.0 & 1.30 \\
Vim-S~\cite{zhu2024vision}         & 95.40 & 0.99 & 82.79 & 3.92 & 25.4 & 1.70 \\
ViT-S~\cite{dosovitskiy2021image}  & 94.00 & 1.05 & 77.12 & 4.83 & 21.7 & 0.97 \\
GLVM~\cite{qin2026neural}          & \textbf{99.40} & 0.20 & \textbf{95.13} & 0.62 & 9.5 & \textbf{0.59} \\
\textbf{Our method}                & 99.13 & \textbf{0.08} & 92.42 & \textbf{0.61} & \textbf{7.2} & 7.28 \\
\hline
\end{tabular}
\end{table}

\subsection{Qualitative Analysis: Branch Activations and Gate Behaviour}

Figure~\ref{fig:branch_gate} reads one correctly classified VERA Palm Vein test image through the last Topology-Aware Block, using forward hooks that leave the forward pass unchanged and Grad-CAM for the predicted class. Each feature panel shows the channel-wise mean absolute activation, min-max normalised to $[0,1]$ and bilinearly resampled to the input grid; the gate panels show the channel mean of $g$ and $g'$, which lie in $[0,1]$ by construction; the curve averages $g$ and $g'$ over all channels, positions, and the $660$ test images, one point per block. The local branch concentrates on fine vessel texture, the structure branch highlights the skeleton of the vascular tree, and the global branch returns a smoother long-range response, so the three encode complementary cues, and the fused maps $L(z)$ and $F(z)$ keep that structural detail rather than washing it out. The gates re-weight the streams from block to block rather than settling on a fixed mixture, and the global branch stays active throughout the body: taking the mean absolute magnitude of each gated stream and normalising to sum to one, the local, structural, and global streams contribute $23.8\%$, $30.8\%$, and $45.4\%$ of the response over the test split.

\FloatBarrier
\section{Conclusion}

In this paper, we presented a topology-aware global-local backbone for palm-vein recognition. The network combines multi-scale local features, a structure-guided directional stream with a fixed Sobel-magnitude edge prior, and a four-direction state-space global pathway within stacked Topology-Aware Blocks, and a staged gated fusion integrates local, structural, and global representations in that order. On HKPU-NIR it achieved $99.13\%$ top-1 accuracy and $0.08\%$ EER with $7.2$~M parameters, and on VERA Palm Vein $92.42\%$ accuracy and $0.61\%$ EER, the lowest EER of every compared backbone on both datasets at the smallest parameter count, while GLVM achieved the highest top-1 identification accuracy. Branch visualisations indicate that the local, structural, and global pathways capture complementary vascular information. Three limitations remain. The prior is a fixed first-order edge operator, and how it compares with a raw-pixel Sobel response, a learnable edge extractor, or a multi-channel or layer-wise re-derived prior is not settled here. The comparison against the four baselines shares their data partition and evaluation protocol but is not a re-run under one training pipeline. The forward pass costs $7.28$~G FLOPs, an order of magnitude above GLVM, which restricts the design to settings where verification reliability outweighs inference budget. Future work will therefore address learnable and connectivity-aware priors, a leaner global operator, baselines retrained in a common pipeline, and extension to other vein modalities and multi-spectral biometric fusion.

\printbibliography

@article{sundararajan2018deep,
  title={Deep learning for biometrics: A survey},
  author={Sundararajan, Kalaivani and Woodard, Damon L},
  journal={ACM Comput. Surv.},
  volume={51},
  number={3},
  pages={1--34},
  year={2018},
  publisher={ACM New York, NY, USA}
}

@article{hong2017convolutional,
  title={Convolutional neural network-based finger-vein recognition using NIR image sensors},
  author={Hong, Hyung Gil and Lee, Min Beom and Park, Kang Ryoung},
  journal={Sensors},
  volume={17},
  number={6},
  pages={1297},
  year={2017},
  publisher={MDPI}
}

@article{das2018convolutional,
  title={Convolutional neural network for finger-vein-based biometric identification},
  author={Das, Rig and Piciucco, Emanuela and Maiorana, Emanuele and Campisi, Patrizio},
  journal={IEEE Trans. Inf. Forensics Security},
  volume={14},
  number={2},
  pages={360--373},
  year={2018},
  publisher={IEEE}
}

@article{qin2017deep,
  title={Deep representation-based feature extraction and recovering for finger-vein verification},
  author={Qin, Huafeng and El-Yacoubi, Mounim A},
  journal={IEEE Trans. Inf. Forensics Security},
  volume={12},
  number={8},
  pages={1816--1829},
  year={2017},
  publisher={IEEE}
}

@inproceedings{he2016deep,
  title={Deep residual learning for image recognition},
  author={He, Kaiming and Zhang, Xiangyu and Ren, Shaoqing and Sun, Jian},
  booktitle={Proc. IEEE Conf. Comput. Vis. Pattern Recognit. (CVPR)},
  pages={770--778},
  year={2016}
}

@inproceedings{woo2018cbam,
  title={Cbam: Convolutional block attention module},
  author={Woo, Sanghyun and Park, Jongchan and Lee, Joon-Young and Kweon, In So},
  booktitle={Proc. Eur. Conf. Comput. Vis. (ECCV)},
  pages={3--19},
  year={2018}
}

@article{vaswani2017attention,
  title={Attention is all you need},
  author={Vaswani, Ashish and Shazeer, Noam and Parmar, Niki and Uszkoreit, Jakob and Jones, Llion and Gomez, Aidan N and Kaiser, {\L}ukasz and Polosukhin, Illia},
  journal={Proc. Adv. Neural Inf. Process. Syst. (NeurIPS)},
  volume={30},
  year={2017}
}

@article{gu2023mamba,
  title={Mamba: Linear-time sequence modeling with selective state spaces},
  author={Gu, Albert and Dao, Tri},
  journal={arXiv preprint arXiv:2312.00752},
  year={2023}
}

@article{zhu2024vision,
  title={Vision mamba: Efficient visual representation learning with bidirectional state space model},
  author={Zhu, Lianghui and Liao, Bencheng and Zhang, Qian and Wang, Xinlong and Liu, Wenyu and Wang, Xinggang},
  journal={arXiv preprint arXiv:2401.09417},
  year={2024}
}

@article{liu2024vmamba,
  title={Vmamba: Visual state space model},
  author={Liu, Yue and Tian, Yunjie and Zhao, Yuzhong and Yu, Hongtian and Xie, Lingxi and Wang, Yaowei and Ye, Qixiang and Jiao, Jianbin and Liu, Yunfan},
  journal={Proc. Adv. Neural Inf. Process. Syst. (NeurIPS)},
  volume={37},
  pages={103031--103063},
  year={2024}
}

@inproceedings{marattukalam2020segmentation,
  title={Segmentation of palm vein images using U-Net},
  author={Marattukalam, Felix and Abdulla, Waleed H},
  booktitle={Proc. Asia-Pacific Signal Inf. Process. Assoc. Annu. Summit Conf. (APSIPA ASC)},
  pages={64--70},
  year={2020},
  organization={IEEE}
}

@inproceedings{shit2021cldice,
  title={clDice-a novel topology-preserving loss function for tubular structure segmentation},
  author={Shit, Suprosanna and Paetzold, Johannes C and Sekuboyina, Anjany and Ezhov, Ivan and Unger, Alexander and Zhylka, Andrey and Pluim, Josien PW and Bauer, Ulrich and Menze, Bjoern H},
  booktitle={Proc. IEEE/CVF Conf. Comput. Vis. Pattern Recognit. (CVPR)},
  pages={16560--16569},
  year={2021}
}

@inproceedings{tome2015palm,
  title={Palm vein database and experimental framework for reproducible research},
  author={Tome, Pedro and Marcel, S{\'e}bastien},
  booktitle={Proc. Int. Conf. Biometrics Special Interest Group (BIOSIG)},
  pages={1--7},
  year={2015},
  organization={IEEE}
}

@article{zhang2010online,
  author  = {Zhang, David and Guo, Zhenhua and Lu, Guangming and Zhang, Lei and Zuo, Wangmeng},
  title   = {An Online System of Multispectral Palmprint Verification},
  journal = {IEEE Trans. Instrum. Meas.},
  volume  = {59},
  number  = {2},
  pages   = {480--490},
  year    = {2010},
}

@inproceedings{dosovitskiy2021image,
  title={An image is worth 16x16 words: Transformers for image recognition at scale},
  author={Dosovitskiy, Alexey and Beyer, Lucas and Kolesnikov, Alexander and Weissenborn, Dirk and Zhai, Xiaohua and Unterthiner, Thomas and Dehghani, Mostafa and Minderer, Matthias and Heigold, Georg and Gelly, Sylvain and Uszkoreit, Jakob and Houlsby, Neil},
  booktitle={Proc. Int. Conf. Learn. Represent. (ICLR)},
  year={2021}
}

@article{qin2026neural,
  title={Neural Architecture Search-Based Global--Local Vision Mamba for Palm-Vein Recognition},
  author={Qin, Huafeng and Fu, Yuming and Chen, Jing and El-Yacoubi, Mounim A and Gao, Xinbo and Xi, Feng},
  journal={IEEE Trans. Inf. Forensics Security},
  volume={21},
  pages={3766--3780},
  year={2026},
  publisher={IEEE}
}

\end{document}